\documentclass[letterpaper]{article} 
\usepackage{aaai2026}  
\usepackage{times}  
\usepackage{helvet}  
\usepackage{courier}  
\usepackage[hyphens]{url}  
\usepackage{graphicx} 
\usepackage{natbib}  
\usepackage{caption} 
\usepackage{algorithm}
\usepackage{algorithmic}
\usepackage{appendix}
\usepackage{amsmath}

\usepackage{booktabs}
\usepackage{multirow}

\usepackage{mathtools}
\usepackage{newfloat}
\usepackage{listings}
\DeclareCaptionStyle{ruled}{labelfont=normalfont,labelsep=colon,strut=off} 
\floatstyle{ruled}
\newfloat{listing}{tb}{lst}{}
\floatname{listing}{Listing}
\title{LeanRAG: Knowledge-Graph-Based Generation with Semantic Aggregation and Hierarchical Retrieval}
\author{
    Yaoze Zhang\textsuperscript{\rm 1,2}\equalcontrib, Rong Wu\textsuperscript{\rm 1,3}\equalcontrib, Pinlong Cai\textsuperscript{\rm 1}\thanks{Corresponding author}, Xiaoman Wang\textsuperscript{\rm 4}, Guohang Yan\textsuperscript{\rm 1}\\ Song Mao\textsuperscript{\rm 1}, Ding Wang\textsuperscript{\rm 1}, Botian Shi\textsuperscript{\rm 1}
}
\nocopyright
\affiliations{
    \textsuperscript{\rm 1}Shanghai Artificial Intelligence Laboratory \\ 
    \textsuperscript{\rm 2}University of Shanghai for Science and Technology \\
     \textsuperscript{\rm 3}Zhejiang University \\
      \textsuperscript{\rm 4}East China Normal University
       
}

\usepackage{bibentry}

\usepackage[table]{xcolor}

\begin{document}

\maketitle

\begin{abstract}
Retrieval-Augmented Generation (RAG) plays a crucial role in grounding Large Language Models by leveraging external knowledge, whereas the effectiveness is often compromised by the retrieval of contextually flawed or incomplete information. To address this, knowledge graph-based RAG methods have evolved towards hierarchical structures, organizing knowledge into multi-level summaries. However, these approaches still suffer from two critical, unaddressed challenges: high-level conceptual summaries exist as disconnected ``semantic islands'', lacking the explicit relations needed for cross-community reasoning; and the retrieval process itself remains structurally unaware, often degenerating into an inefficient flat search that fails to exploit the graph's rich topology. To overcome these limitations, we introduce LeanRAG, a framework that features a deeply collaborative design combining knowledge aggregation and retrieval strategies. LeanRAG first employs a novel semantic aggregation algorithm that forms entity clusters and constructs new explicit relations among aggregation-level summaries, creating a fully navigable semantic network. Then, a bottom-up, structure-guided retrieval strategy anchors queries to the most relevant fine-grained entities and then systematically traverses the graph's semantic pathways to gather concise yet contextually comprehensive evidence sets. The LeanRAG can mitigate the substantial overhead associated with path retrieval on graphs and minimize redundant information retrieval. Extensive experiments on four challenging QA benchmarks with different domains demonstrate that LeanRAG significantly outperforms existing methods in response quality while reducing 46\% retrieval redundancy. Our code is available at: \url{https://github.com/RaZzzyz/LeanRAG}.

\end{abstract}

\section{Introduction}

Large Language Models (LLMs) have demonstrated remarkable capabilities in natural language understanding and generation. Yet their effectiveness is often undermined by their static internal knowledge, leading to factual inaccuracies and hallucinations \cite{huang2025survey, Li2024TheDAA}. Retrieval-Augmented Generation (RAG) was introduced as a potential solution, dynamically grounding LLMs in external, up-to-date information \cite{Gao2023RetrievalAugmentedGFA}. However, the effectiveness of naive RAG approaches is frequently compromised. The retrieved text chunks often lack precise alignment with the user's true intent, and the reliance on embedding-based similarity alone is often insufficient to capture the deep semantic relevance required for complex reasoning, resulting in responses that are either incomplete or contextually flawed \cite{Zhao2024RetrievalAugmentedGFA, ICLR2025_2ea06b52}.

\begin{figure*}[!t]
    \centering
    \includegraphics[width=0.82\textwidth]{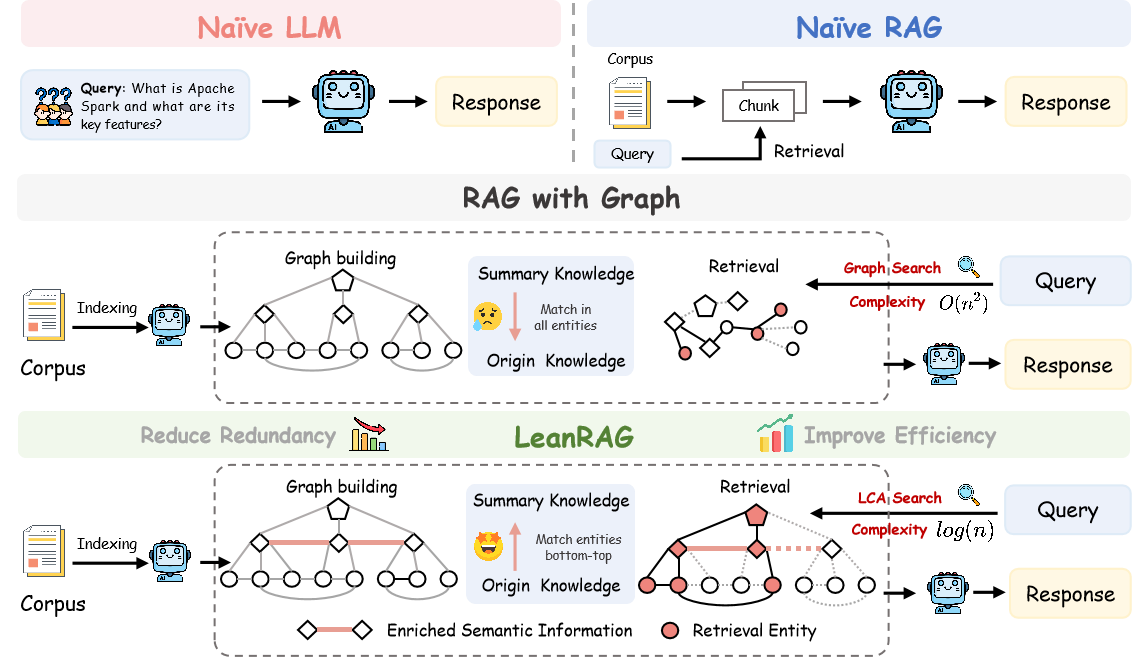}
    \caption{Comparison of typical LLM retrieval-augmented generation frameworks.}
    \label{fig:teaser}
\end{figure*}

To overcome the limitations of unstructured retrieval, researchers have increasingly explored knowledge graph-based RAG methods. Initial efforts, such as GraphRAG \cite{graphrag}, successfully organized documents into community-based knowledge graphs, which helped preserve local context better than disconnected text chunks. However, these methods often generated large, coarse-grained communities, leading to significant information redundancy during retrieval. Subsequently, more advanced works like HiRAG \cite{hirag} refined this paradigm by introducing hierarchical structures, clustering entities into multi-level summaries. This represented a significant step forward in organizing knowledge. Despite this progress, our analysis reveals that two critical challenges remain unaddressed currently (as Figure~\ref{fig:teaser} shows). First, the high-level summary nodes in these hierarchies exist as ``semantic islands''. They lack explicit relational connections between each other, making it hard to reason across different conceptual communities within the knowledge base. Second, the retrieval process itself remains structurally unaware, often degenerating into a simple semantic search over a flattened list of nodes, failing to exploit the rich topological information encoded in the graph. This leads to a retrieval process that is both inefficient and imprecise.

To address these challenges, we propose LeanRAG, a novel retrieval-augmented generation framework that synergistically integrates deeply collaborative knowledge structuring with a lean, structure-guided retrieval strategy. At its core, LeanRAG introduces a semantic aggregation algorithm that constructs a hierarchical knowledge graph by organizing retrieved entities into semantically coherent clusters. Its key innovation lies not only in clustering entities based on semantic similarity but also in automatically inferring explicit inter-cluster summary relations, leveraging the underlying knowledge’s contextual and relational semantics to establish higher-order abstractions. This process transforms fragmented, isolated hierarchies into a unified, fully navigable semantic network, where both fine-grained details and abstracted knowledge are seamlessly interconnected.

Building upon this enriched structure, LeanRAG employs a bottom-up, structure-aware retrieval mechanism that strategically navigates the graph to maximize relevance while minimizing redundancy. The retrieval process begins by anchoring the query to the most contextually pertinent fine-grained entities at the leaf level. It then systematically traverses relational pathways across both the original entity layer and the derived summary layer, propagating evidence upward through the hierarchy. This dual-level traversal ensures that the retrieved evidence set is not only concise and focused but also contextually comprehensive, capturing both specific details and broader conceptual relations essential for accurate and coherent generation.

Our primary contributions can be summarized as follows:

\begin{itemize}

\item A novel semantic aggregation algorithm designed for superior knowledge condensation. This method constructs a multi-resolution knowledge map by modeling and building new relational edges between summary-level conceptual nodes, effectively preserving both fine-grained facts and high-level thematic connections within a single, coherent structure.

\item The introduction of a bottom-up entity retrieval strategy to mitigate information redundancy. By initiating retrieval from high-relevance ``anchor'' nodes and expanding context strictly along relevant semantic pathways, this strategy yields a precise and compact evidence subgraph for LLMs.

\item We demonstrate through extensive experiments that LeanRAG achieves a new state-of-the-art on multiple challenging QA tasks, significantly outperforming existing methods in both response performance and efficiency.

\end{itemize}

\section{Related Work}

\subsection{Retrieval-Augmented Generation}
Retrieval-Augmented Generation was introduced as a powerful paradigm to mitigate the intrinsic knowledge limitations of LLMs by grounding them in external information \cite{naiverag}. The standard RAG framework operates by retrieving relevant text chunks from a corpus and providing them as context to an LLM for answer generation \cite{wang2024searching}. While effective, this approach is fundamentally constrained by the ``chunking dilemma'': small, fine-grained chunks risk losing critical context, whereas larger chunks often introduce significant noise and dilute the LLM's focus \cite{Tonellotto2024ThePOA}.

Substantial research has been dedicated to overcoming this limitation. One line of work improves the retriever itself, evolving from sparse methods like BM25 \cite{robertson2009probabilistic} to dense models such as DPR \cite{karpukhin2020dense} and Contriever \cite{izacard2021unsupervised}, which better capture semantic relevance. Another focuses on indexing and organizing source documents \cite{jiang2023active}, with recent methods creating hierarchical summaries of text chunks to enable multi-level retrieval. For example, RAPTOR builds a tree of recursively summarized clusters, allowing retrieval of fine-grained details and high-level summaries \cite{sarthi2024raptor}. However, these approaches still treat knowledge as linear or simple hierarchical structures and do not explicitly model complex, non-hierarchical relations between entities and concepts, limiting their ability to answer queries requiring reasoning over such connections—motivating KG-based RAG methods.

\subsection{Knowledge Graph Based Retrieval-Augmented Generation}

To better capture the relational nature of information, KG-based RAG has emerged as a prominent research direction. By representing knowledge as a graph of entities and relations, these methods aim to provide a more structured and semantically rich context for the LLM \cite{Peng2024GraphRGA}. Early approaches in this domain focused on leveraging graph structures for improved retrieval. For instance, GraphRAG \cite{graphrag} organizes documents into community-based KGs to preserve local context, while other methods like FastGraphRAG utilize graph-centrality metrics such as PageRank \cite{pagerank} to prioritize more important nodes during retrieval. This subgraph retrieval approach has also proven effective in industrial applications like customer service, where KGs are constructed from historical support tickets to provide structured context \cite{Xu2024RetrievalAugmentedGWA}. These methods marked a significant step forward by imposing a macro-structure onto the knowledge base, moving beyond disconnected text chunks.

Recognizing the need for more fine-grained control and abstraction, subsequent works have explored more sophisticated hierarchical structures. HiRAG, the current state-of-the-art, clusters entities to form multi-level summaries \cite{hirag}, while LightRAG  \cite{lightrag} proposes a dual-level framework to balance global and local information retrieval. While these hierarchical methods have progressively improved retrieval quality, a critical gap persists in how the constructed graph structures are leveraged at query time. The retrieval process is often decoupled from the indexing structure; for instance, an initial search may be performed over a ``flattened'' list of all nodes, rather than being directly guided by the indexed community or hierarchical relations. This decoupling means the rich structural information is primarily used for post-retrieval context expansion, rather than for guiding the initial, crucial step of identifying relevant information. This can limit performance on complex queries where the relations between entities are paramount, highlighting the need for a new paradigm where the retrieval process is natively co-designed with the knowledge structure.

\section{Preliminary}
In this section, we will introduce and give a formal definition of a RAG system with a specific knowledge graph.

Given a rich knowledge graph with the description of vertexes and relations $\mathcal G = (V, R, D_{\text(ver)}, D_{\text(rel)})$, where $V$ and $R$ denote the set of entities and relations, $D_{\text(ver)}$ represents the collection of entity descriptions and $D_{\text(rel)}$ represents the collection of relationship descriptions. The goal of KG-based RAG is to leverage existing information to build a query-relevant sub-graph that helps LLMs generate high-quality responses. Given a query $q$, the searching process can be formulated as:

\begin{equation}
\label{e1}
    \tilde{V} = \text{Top-n}_{v \in V} (\text{Sim}(q, d_v))
\end{equation} 
where $Sim(\cdot, \cdot)$ is the embedding similarity metric function, and n is the choice number of similarity entities. Based on the metric, $\tilde{V}$ contains the top n entities. Then we can search the relational paths $L$ between nodes $ v \in \tilde{V}$. All relations $r$ that constitute the path $L$ belong to the relation set $R$.

\begin{equation}
    L =  \displaystyle\bigcup\limits_{ x,y \in \tilde{\mathcal{V}}} Path(x, y) = (r_1, r_2, \dots)
\end{equation}
By leveraging $\tilde{V}$ and $L$, the sub-graph $\tilde{\mathcal G}$ is constructed to support RAG systems with focused, query-relevant, and semantically enriched knowledge retrieval.

\begin{figure*}[!t]
    \centering
    \includegraphics[width=0.94\linewidth]{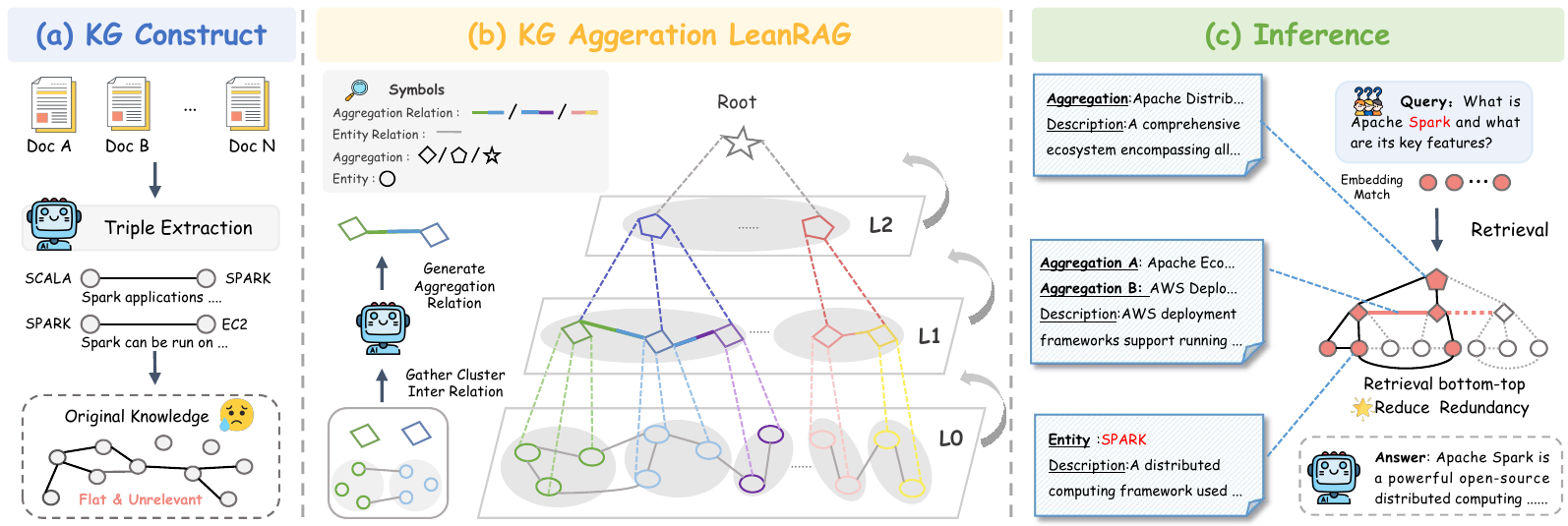}
    \caption{Overview of the LeanRAG framework.}
    \label{fig:framework}
\end{figure*}

\section{Method}
The performance of a generic KG-augmented retrieval framework is fundamentally determined by the structural and semantic quality of the underlying knowledge graph $\mathcal{G}$, as well as the precision and efficiency of the retrieval strategy. To address the limitations of a flat graph structure and naive path search strategy, we introduce \textbf{LeanRAG}, a framework built on the principle of tightly \textbf{co-designing} its aggregation and retrieval processes. As illustrated in Figure~\ref{fig:framework}, LeanRAG consists of two core innovations: (1) a \textbf{Hierarchical Graph Aggregation} method that recursively builds a multi-level, navigable semantic network from the base KG; and (2) a \textbf{Structured Retrieval} strategy that leverages this hierarchy via Lowest Common Ancestor (LCA) path search approach to construct a compact and coherent context.

\subsection{Hierarchical Knowledge Graph Aggregation}
\label{sec:aggregation}

The foundation of LeanRAG is the transformation of a flat knowledge graph $\mathcal{G}_0$ into a multi-level, semantically rich hierarchy $\mathcal{H}$. This hierarchy allows for retrieval at varying levels of abstraction. We construct this hierarchy, denoted as $\mathcal{H} = \{\mathcal{G}_0, \mathcal{G}_1, \dots, \mathcal{G}_k\}$, in a bottom-up, layer-by-layer fashion. Each layer $\mathcal{G}_i = (V_i, R_i, D_{{\text(ver)}_i}, D_{{\text(rel)}_i})$ represents a more abstract view of the layer below it, $\mathcal{G}_{i-1}$. The core of this construction lies in a recursive aggregation process that clusters nodes based on semantic similarity and then intelligently generates new, more abstract entities and relations to form the next layer.

\subsubsection{Recursive Semantic Clustering.}

Given a knowledge graph layer $\mathcal{G}_{i-1}$, the first step is to identify groups of semantically related entities that can be abstracted into a single, higher-level concept. We leverage the rich descriptive text $d_v \in D_{{\text(ver)}_{i-1}}$ associated with each entity $v \in V_{i-1}$ for this purpose. Following recent works in clustering text representation  \cite{sarthi2024raptor}, we employ a two-step process:

\begin{enumerate}
    \item \textbf{Semantic Embedding:} We first encode the textual description of each entity into a dense vector representation using a pre-trained embedding model $\Phi(\cdot)$. This yields a set of embeddings for the entire KG layer:

    \begin{equation}
        \mathbf{E}_{i-1} = \{ \Phi(d_v) \mid v \in V_{i-1} \}
    \end{equation}

    \item \textbf{Gaussian Mixture Clustering:} We then apply a Gaussian Mixture Model (GMM) \cite{reynolds2015gaussian}  to the set of embeddings $\mathbf{E}_{i-1}$. The GMM partitions the entities $V_{i-1}$ into $m$ disjoint clusters $\mathcal{C}_{i-1} = \{C_1, C_2, \dots, C_m\}$, where each cluster $C_j\ (j\in [1,m])$ contains entities that are semantically similar in the embedding space.
\end{enumerate}

This clustering provides a principled grouping of fine-grained entities, setting the stage for conceptual abstraction.

\subsubsection{Generation of Aggregated Entities and Relations.}

A key limitation of prior hierarchical methods is that they often only cluster entities, losing the rich relational information in the process. LeanRAG overcomes this by using LLMs to intelligently generate both new entities and new relations for the subsequent layer $\mathcal{G}_i$.

\paragraph{\textit{Aggregated Entity Generation.}}
For each cluster $C_j \in \mathcal{C}_{i-1}$, we generate a single, more abstract aggregated entity $\alpha_j$ that represents the cluster's collective semantics. This abstraction is achieved via a generation function $\mathcal{F}_{\text{entity}}$, which synthesizes a new concept by considering both the entities within the cluster and the relations that exist among them. Let ${R}_{C_j}$ be the set of relations in $\mathcal{G}_{i-1}$ among entities within cluster $C_j$.

\begin{equation}
    (\alpha_j, d_{\alpha_j}) = \mathcal{F}_{\text{entity}}(C_j, R_{C_j})
\end{equation}
The new entity set $V_i = \{\alpha_j\}_{j=1}^m$  and their associated descriptions $D_{V_i}\{d_{\alpha_j}\}_{j=1}^m$   are defined as the parent nodes of $\{C_1, C_2, \dots, C_m\}$ in the hierarchy, i.e., the nodes located at the immediate higher level in the hierarchical structure.

In practice, the generation function $\mathcal{F}_{\text{entity}}$ is implemented by LLMs guided by a carefully designed prompt $\mathcal{P}_{\text{entity}}$. We prompt LLMs to produce a concise name for the new entity $\alpha_j$ and a comprehensive description $d_{\alpha_j}$ that summarizes its components. Each entity $v \in C_j$ is then linked to its new parent entity $\alpha_j$, forming the parent-child connections in the hierarchy.

\paragraph{\textit{Aggregated Relation Generation.}}
To prevent the formation of ``semantic islands'' at higher layers, we explicitly create new relations between the aggregated entities in $V_i$. This ensures that the graph remains connected and navigable at all levels of abstraction. For any pair of aggregated entities $(\alpha_j, \alpha_k)$, we confirm the inter-cluster relations $R_{<C_j,C_k>}$ that contains the relations between nodes that belong to the $C_j$ and $C_k$, respectively. Then, we constitute the inter-cluster aggregated relation $r_{<C_j,C_k>}$ by $R_{<C_j,C_k>}$. This paper defines the number of $R_{<C_j,C_k>}$ as the connectivity strength, $\lambda_{j, k}$. If $\lambda_{j, k}$ exceeds a dynamically defined threshold $\tau$, we infer that a meaningful high-level relationship exists, which is summarized by the LLM-driven function $\mathcal{F}_{\text{rel}}$. Otherwise, the inter-cluster aggregated relation is simply regarded as the text concatenation of $R_{<C_j,C_k>}$.
\begin{equation}
\label{eq:relation_generation}
\hspace{-0.1cm}
r_{<\alpha_j, \alpha_k>}= 
\begin{cases} 
    \mathcal{F}_{\text{\text(rel)}}(\alpha_j, \alpha_k, R_{<C_j,C_k>}), & \text{if } \lambda_{j, k} > \tau \\ 
    {Concate}(R_{<C_j,C_k>}),   & \text{otherwise}
\end{cases}
\end{equation}

In practice, the generation function $\mathcal{F}_{\text{rel}}$ is implemented by LLMs guided by a specific prompt $\mathcal{P}_{\text{rel}}$. 

The threshold $\tau$ is a data-dependent hyper-parameter that may vary with the layer index to reflect the knowledge graph’s density at different abstraction levels, ensuring only salient, well-supported relations are propagated.

By recursively applying this process of clustering and generation, we construct a rich, multi-layered KG where each layer provides a progressively more abstract, yet semantically coherent, view of the original information.

\subsection{Structured Retrieval via Lowest Common Ancestor}
\label{sec:retrieval}

The hierarchical knowledge graph $\mathcal{H}$ enables a retrieval strategy that is fundamentally more structured and efficient than searching over a flat graph. Our approach moves beyond simple similarity-based retrieval by leveraging the graph's topology to construct a compact and contextually coherent subgraph. This process consists of two main phases: initial entity anchoring at the base layer, followed by a structured traversal of the hierarchy to gather context.

\subsubsection{Initial Entity Anchoring.}

Given a user query $q$, the first step is to ground the query in the most specific, fine-grained facts available. We achieve this by performing a dense retrieval search exclusively over the entities of the original graph, including the initial entities, that is, the base-layer graph $\mathcal{G}_0$. We identify the top n entities whose textual descriptions are most semantically similar to the query:
\begin{equation}
    V_{\text{seed}} = \text{Top-n}_{v \in V_0} (\text{sim}(q, d_v))
\end{equation}

This set of ``seed entities'', $V_{\text{seed}}$, serves as the starting point for structured traversal, ensuring our retrieval process is anchored in the most relevant parts of the knowledge base.

\subsubsection{Contextualization via LCA Path Traversal.}

Graph retrieval methods in the prior KG-based RAG would typically find all paths between entities in $V_{\text{seed}}$ on the flat graph $\mathcal{G}_0$. This approach often retrieves a large number of intermediate nodes that add noise and redundancy. In contrast, LeanRAG utilizes the entire hierarchy $\mathcal{H}$ to define a much more focused and meaningful context. Our core idea is to construct a minimal subgraph that connects the seed entities through their most immediate shared concepts in the hierarchy. We achieve this using the principle of the LCA. For two seed entities in $V_{\text{seed}}$, their lowest common ancestor (LCA) $v_{\text{lca}}$ is defined as the common ancestor with the minimum depth in the hierarchy $\mathcal{H}$ among all their ancestors. This ensures that the combined path length from the two seed entities to $v_{\text{lca}}$ is minimized to avoid information redundancy.

The retrieval path $\mathcal{P}_{\text{lca}}$ is then defined as the union of all shortest paths in the hierarchy from each seed entity $v \in V_{\text{seed}}$ to the common ancestor $v_{\text{lca}}$:
\begin{equation}
    \mathcal{P}_{\text{lca}}(V_{\text{seed}}, \mathcal{H}) = \bigcup_{v \in V_{\text{seed}}} \text{ShortestPath}_{\mathcal{H}}(v, v_{\text{lca}})
\end{equation}
where $\text{ShortestPath}_{\mathcal{H}}(\cdot,\cdot)$ denotes the shortest path between two nodes within the hierarchical graph $\mathcal{H}$. Since our hierarchy is tree-like, this path consists of the direct chain of from child nodes to parent nodes. Finally, the retrieved subgraph for RAG context $\mathcal{G}_{\text{ret}}$ is composed of all entities and relations that lie on these LCA paths:
\begin{equation}
\mathcal{G}_{\text{ret}} = (V_{\text{ret}}, R_{\text{ret}} )
\end{equation}
\begin{equation}
    V_{\text{ret}} = \{ v \mid v \in \mathcal{P}_{\text{lca}} \}
\end{equation}
\begin{equation}
    R_{\text{ret}} = R_{\text{lca}} \cup R_{\text{inter-cluster}} 
\end{equation}
where $R_{\text{lca}}$ contains the relations within the retrieval path $\mathcal{P}_{\text{lca}}$ and $R_{\text{inter-cluster}}$ contains the inter-cluster relations between aggregation entities that are in the same level in the hierarchical knowledge graph. For example, $r_{<\alpha_j, \alpha_k>} \in R_{\text{inter-cluster}}$, where $\alpha_j \in \mathcal{G}_i$ and $\alpha_k \in \mathcal{G}_i$.

This LCA-based traversal strategy ensures that the retrieved context is not just a collection of relevant entities, but a connected, coherent narrative structure, spanning from specific facts to their shared abstract concepts. This significantly reduces information redundancy and provides a much richer, more structured context to the final LLM generator. Furthermore, we return the original chunks from which the entities were sourced as supporting evidence. The illustration of this process is provided in Figure~\ref{fig:framework}.

\section{Experiments}

In our experiments, we aim to answer the following research questions:
\begin{itemize}
    \item RQ1: How does LeanRAG's \textbf{QA performance} compare against state-of-the-art baselines across diverse domains?
    \item RQ2: Does LeanRAG's retrieval strategy \textbf{reduce information redundancy} while improving generation quality?
    \item RQ3: To what extent does the explicit generation of \textbf{relations between aggregated entities} contribute to the quality of the response?
    \item RQ4: Is the structured knowledge retrieved from the graph sufficient for high-quality generation, or is the inclusion of the entities \textbf{original textual context essential}?
    
\end{itemize}

\subsubsection{Baselines.}
To evaluate the performance of LeanRAG, we compare it against a comprehensive suite of representative and state-of-the-art KG-based RAG methods. The selected baselines include:
\begin{itemize}
    \item \textbf{NaiveRAG} \cite{naiverag}: The foundational RAG approach, which retrieves semantically similar text chunks from a document corpus.
    \item \textbf{GraphRAG} \cite{graphrag}: A prominent KG-based method that organizes knowledge into communities. We utilize its local search mode, as the global mode has significant computational overhead and does not leverage local entity context.
    \item \textbf{LightRAG} \cite{lightrag}: Uses a dual-level retrieval framework based on a KG-based text indexing paradigm.
    \item \textbf{KAG} \cite{kag}: A pipeline that aligns LLM generation with structured KG reasoning through mutual knowledge-text indexing and logic-form guidance.
    \item \textbf{FastGraphRAG}: An enhancement of graph retrieval that uses the PageRank algorithm \cite{pagerank} to prioritize nodes of higher importance.
    \item \textbf{HiRAG} \cite{hirag}: The current state-of-the-art, which introduces hierarchical structures by clustering entities into multi-level summaries.
\end{itemize}

\subsubsection{Datasets and Evaluation Metrics.}
We used four datasets from the UltraDomain benchmark \cite{qian2024memorag}, which is designed to evaluate RAG systems across diverse applications, focusing on long context tasks and high-level queries in specialized domains. We used Mix, CS, Legal, and Agriculture datasets following the prior work \cite{lightrag}.

\paragraph{\textit{Evaluation Metrics.}}
To provide a multi-faceted and in-depth analysis of system performance, we evaluate the generated answers along four crucial dimensions, following the prior work \cite{hirag}:


\begin{itemize}
    \item \textbf{Comprehensiveness:} Measures how thoroughly the answer addresses the user's query.

    \item \textbf{Empowerment:} Evaluates the answer's practical utility and its ability to provide actionable information.

    \item \textbf{Diversity:} Assesses the breadth of information and perspectives presented in the answer.

    \item \textbf{Overall:} Provides a single, holistic quality score to measure how the answer performs overall, considering comprehensiveness, empowerment, diversity, and any other relevant factors.
\end{itemize}

Following recent best practices in automated evaluation, we employ powerful LLMs as judges to score the outputs of all methods on the 1 to 10 scale defined by our metrics. In order to directly reflect the quality of the answers, we will also use LLM to directly evaluate the two answers to obtain their win rates. Specifically, we use DeepSeek-V3 \cite{liu2024deepseek} as our evaluators, providing them with carefully designed prompts to ensure consistent and unbiased scoring, and each query and answer is scored 5 times.

\subsubsection{Implementation Details.}
Across all experiments, we use DeepSeek-V3 as an LLM generator for all models to ensure a fair comparison. The text embedding for retrieval is computed using BGE-M3 \cite{bge-m3}. The number of clusters for the GMM and other key hyperparameters are tuned on a held-out validation set. All main experiments were conducted by leveraging commercial API services. For our main experiments, we utilized the Deepseek-V3 model as the backbone for all models, following the prior work \cite{hirag}, ensuring a fair comparison. In addition, in order to evaluate RQ2 efficiently, we reproduced the baseline methods on the Qwen3-14b \cite{yang2025qwen3technicalreport} model to evaluate the redundancy between LeanRAG and other methods. All implementation details are provided in the extended version \cite{zhang2025leanragknowledgegraphbasedgenerationsemantic}.

\subsection{Overall Performance Comparison (RQ1)}
To address RQ1 , we compare LeanRAG against all baseline models across four benchmarks, as presented in Table~\ref{exp:rq1}. The experimental results demonstrate that LeanRAG almost outperforms all baselines across the evaluated datasets.

\begin{table*}[t]
\small
\centering

\begin{tabular}{>{\centering\arraybackslash}m{1.2cm}>{\raggedright\arraybackslash}p{2.5cm}>{\centering\arraybackslash}m{1.45cm}>{\centering\arraybackslash}m{1.45cm}>{\centering\arraybackslash}m{1.45cm}>{\centering\arraybackslash}m{1.45cm}>{\centering\arraybackslash}m{1.45cm}>{\centering\arraybackslash}m{1.45cm}>{\centering\arraybackslash}m{1.45cm}}
\toprule
\textbf{Dataset} & \textbf{Metric} $\uparrow$ & \cellcolor{gray!20} \textbf{LeanRAG} & HiRAG & Naive & GraphRAG & LightRAG & FastGraphRAG & KAG \\
\midrule

\multirow{4}{*}{Mix}
& Comprehensiveness & \cellcolor{gray!20} \textbf{8.89$\pm$0.01} & 8.72$\pm$0.02 & 8.20$\pm$0.01 & 8.52$\pm$0.01 & 8.19$\pm$0.02 & 6.56$\pm$0.02 & 7.90$\pm$0.03 \\
& Empowerment       & \cellcolor{gray!20} \textbf{8.16$\pm$0.02} & 7.86$\pm$0.03 & 7.52$\pm$0.03 & 7.73$\pm$0.02 & 7.56$\pm$0.03 & 5.82$\pm$0.03 & 7.41$\pm$0.04 \\
& Diversity         & \cellcolor{gray!20} \textbf{7.73$\pm$0.01} & 7.21$\pm$0.02 & 6.65$\pm$0.03 & 7.04$\pm$0.02 & 6.69$\pm$0.04 & 4.88$\pm$0.03 & 6.42$\pm$0.04 \\
& Overall           & \cellcolor{gray!20} \textbf{8.59$\pm$0.01} & 8.08$\pm$0.02 & 7.47$\pm$0.02 & 7.87$\pm$0.01 & 7.61$\pm$0.04 & 5.76$\pm$0.02 & 7.25$\pm$0.03 \\
\midrule

\multirow{4}{*}{CS}
& Comprehensiveness & \cellcolor{gray!20} 8.92$\pm$0.01 & 8.92$\pm$0.01 & \textbf{8.94$\pm$0.01} & 8.55$\pm$0.02 & 8.76$\pm$0.02 & 6.79$\pm$0.01 & 8.22$\pm$0.02 \\
& Empowerment       & \cellcolor{gray!20} 8.68$\pm$0.02 & 8.66$\pm$0.02 & \textbf{8.69$\pm$0.04} & 8.28$\pm$0.04 & 8.50$\pm$0.04 & 6.67$\pm$0.04 & 8.52$\pm$0.05 \\
& Diversity         & \cellcolor{gray!20} \textbf{7.87$\pm$0.02} & 7.84$\pm$0.02 & 7.79$\pm$0.02 & 7.42$\pm$0.02 & 7.63$\pm$0.04 & 5.45$\pm$0.04 & 7.03$\pm$0.02 \\
& Overall           & \cellcolor{gray!20} \textbf{8.82$\pm$0.02} & 8.77$\pm$0.02 & 8.77$\pm$0.03 & 8.37$\pm$0.04 & 8.59$\pm$0.04 & 6.31$\pm$0.03 & 7.99$\pm$0.03 \\
\midrule

\multirow{4}{*}{Legal}
& Comprehensiveness & \cellcolor{gray!20}8.88$\pm$0.02   & 8.68$\pm$0.02 & 8.85$\pm$0.01 & \textbf{8.95$\pm$0.01} & 8.24$\pm$0.02 & 3.87$\pm$0.02 & 8.41$\pm$0.02 \\
& Empowerment       & \cellcolor{gray!20}\textbf{8.42$\pm$0.03}   & 8.18$\pm$0.06 & 8.28$\pm$0.03 & 8.33$\pm$0.02 & 7.83$\pm$0.05 & 3.53$\pm$0.03 & 8.20$\pm$0.03 \\
& Diversity         & \cellcolor{gray!20}\textbf{7.49$\pm$0.03}   & 7.00$\pm$0.03 & 7.10$\pm$0.04 & 7.47$\pm$0.03 & 6.87$\pm$0.01 & 2.87$\pm$0.02 & 6.71$\pm$0.01 \\
& Overall           & \cellcolor{gray!20}\textbf{8.49$\pm$0.04}   & 8.00$\pm$0.04 & 8.21$\pm$0.03 & 8.44$\pm$0.01 & 7.74$\pm$0.03 & 3.43$\pm$0.02 & 7.83$\pm$0.03 \\
\midrule

\multirow{4}{*}{Agriculture}
& Comprehensiveness & \cellcolor{gray!20} 8.94$\pm$0.06 & \textbf{8.99$\pm$0.00} & 8.85$\pm$0.01 & 8.97$\pm$0.01 & 8.71$\pm$0.01 & 3.28$\pm$0.01 & 8.22$\pm$0.01 \\
& Empowerment       & \cellcolor{gray!20} \textbf{8.66$\pm$0.02} & 8.52$\pm$0.02 & 8.51$\pm$0.03 & 8.52$\pm$0.02 & 8.23$\pm$0.02 & 3.29$\pm$0.05 & 8.33$\pm$0.06 \\
& Diversity         & \cellcolor{gray!20} \textbf{8.06$\pm$0.03} & 7.98$\pm$0.02 & 7.76$\pm$0.06 & 7.95$\pm$0.02 & 7.68$\pm$0.03 & 3.01$\pm$0.03 & 7.07$\pm$0.02 \\
& Overall           & \cellcolor{gray!20} \textbf{8.87$\pm$0.02} & 8.87$\pm$0.03 & 8.69$\pm$0.03 & 8.85$\pm$0.01 & 8.56$\pm$0.02 & 3.17$\pm$0.02 & 7.95$\pm$0.03 \\
\bottomrule

\end{tabular}
\caption{Evaluation scores (1–10 scale) of LeanRAG compared to baseline methods, assessed by an LLM }
\label{exp:rq1}
\end{table*}

From a \textit{Comprehensiveness} perspective, even after removing the information-intensive community structure of traditional KG-based RAG, the aggregation used by LeanRAG still provides sufficient query-related information. Furthermore, \textit{Empowerment} and \textit{Diversity} effectively measure the relevance of the provided information. These indicate that LeanRAG effectively enhances the breadth of information by establishing inter-cluster relations, resulting in optimal performance. In summary, LeanRAG demonstrates state-of-the-art performance on the majority of metrics across four evaluated datasets and achieves highly competitive results on the remaining ones.

\subsection{Analysis of Information Redundancy (RQ2)}

\paragraph{\textit{Experimental Setup.}}
To answer RQ2, we evaluate the information redundancy of different methods. We use the token count of the retrieved context as a metric for redundancy, where a lower token count at a comparable performance level signifies a less redundant context. We re-implemented all baselines with Qwen3-14B-Instruct.

\begin{figure}[t]
    \centering
    \includegraphics[width=0.9\linewidth]{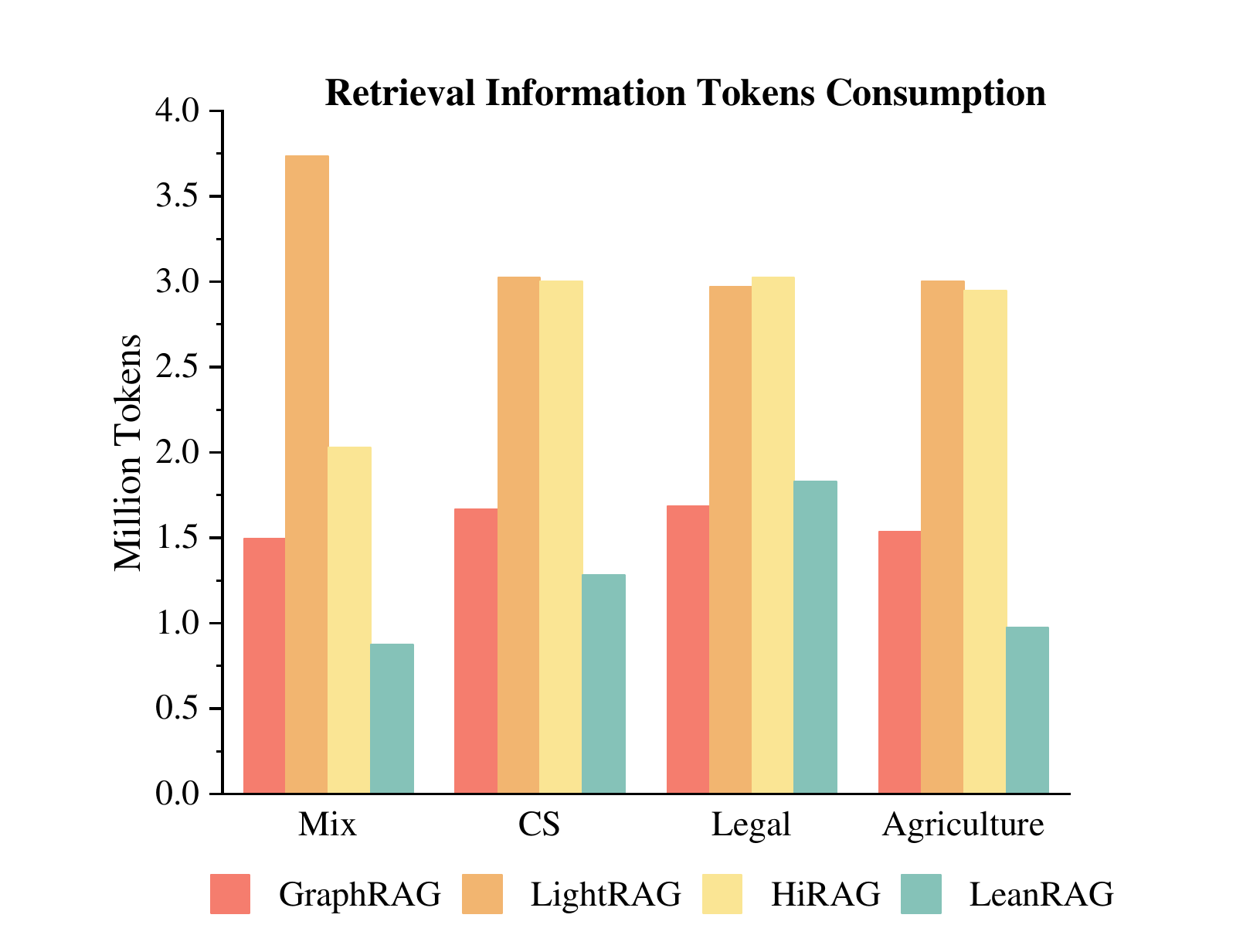}
    \caption{Comparison in retrieval tokens across four datasets}
    \label{fig:retrieval_tokens}
\end{figure}

\paragraph{\textit{Retrieved Context Size.}}
Figure~\ref{fig:retrieval_tokens} shows the number of tokens in the context retrieved by each method. The results indicate that LeanRAG retrieves a substantially more compact context compared to all baselines. On average, its retrieved context is 46\% smaller than baselines. This result can be attributed to our LCA-based traversal strategy, which constructs a focused subgraph by navigating the hierarchy, in contrast to methods that retrieve larger communities.

\subsection{Cluster Relation Effectiveness Analysis (RQ3)}
The core innovation of LeanRAG is not only its use of fine-grained, controllable aggregate entities but also its establishment of paths between them, which creates a fully navigable semantic network for retrieval. This design directly addresses RQ3: whether the inter-cluster relationships, which break the traditional ``semantic islands'' problem, can truly improve retrieval quality. To test this, we conducted experiments on four datasets, comparing the retrieval results of LeanRAG with and without the inclusion of path information. The win rates across four different metrics were then analyzed, with the results summarized in Table~\ref{tab:rq3}.

\begin{table*}[hbtp]
\small

\begin{tabular}{>{\raggedright\arraybackslash}m{2.5cm}>{\raggedright\arraybackslash}p{1.42cm}>{\centering\arraybackslash}m{1.42cm}>{\centering\arraybackslash}m{1.42cm}>{\centering\arraybackslash}m{1.42cm}>{\centering\arraybackslash}m{1.42cm}>{\centering\arraybackslash}m{1.42cm}>{\centering\arraybackslash}m{1.42cm}>{\centering\arraybackslash}m{1.42cm}}
\toprule
\textbf{}    & \multicolumn{2}{c}{\textbf{Mix}} & \multicolumn{2}{c}{\textbf{CS}} & \multicolumn{2}{c}{\textbf{Legal}} & \multicolumn{2}{c}{\textbf{Agriculture}} \\
\midrule
Comprehensiveness& \textbf{51.5}\%& 48.6\%& \textbf{54.5\%}& 45.5\%& \textbf{55.5\%}& 44.5\%& \textbf{54.0\%}& 46.0\%\\
Empowerment& \textbf{55.0\%}& 45.0\%& \textbf{55.5\%}& 44.5\%& \textbf{56.5\%}& 43.5\%& \textbf{59.5\%}& 40.5\%\\
Diversity& \textbf{59.6\%}& 40.4\%& \textbf{66.0\%}&  34.0\%& \textbf{57.0\%}& 43.0\%& \textbf{63.0\%}&37.0\%\\
Overall& \textbf{53.8\%}& 46.2\%& \textbf{58.5\%}& 41.5\%& \textbf{56.5\%}& 43.5\%& \textbf{58.0\%}&42.0\%\\

\bottomrule

\end{tabular}
\caption{Win rates (\%) between LeanRAG and LeanRAG w/o Relation (Left: LeanRAG; Right: w/o Relation)}
\label{tab:rq3}
\end{table*}

The data in Table 3 clearly shows that when relational paths are removed, LeanRAG's retrieval diversity, or the breadth of its information, decreases significantly. This result confirms that establishing relationships between clusters effectively connects isolated entities, thereby enriching the information available for retrieval. Furthermore, by explicitly returning these relationships, the retrieval process is enhanced, leading to a demonstrable improvement in the overall quality of the retrieved answers.




\subsection{Necessity Analysis of Textual Context (RQ4)}
\paragraph{\textit{Motivation and Setup.}}
To answer RQ4, we investigate the role of the original unstructured text chunks in our framework. While the graph structure serves as an effective retrieval guide, it is crucial to assess whether structured information alone suffices for the generator or if the source text remains essential. To this end, we conduct an ablation study using a variant of our model, denoted as \textbf{LeanRAG w/o Context}. This variant follows the same hierarchical retrieval process, but the final context provided to the LLM generator includes only the names and descriptions of the retrieved graph entities, excluding the original text chunks linked to the base-level entities. We then compare its performance with that of the full LeanRAG model.

\paragraph{\textit{Results and Analysis.}}
The results of this comparison are presented in Table~\ref{tab:rq4_context_necessity}. Across all four datasets and nearly every evaluation metric, the performance of LeanRAG drops significantly when the original textual context is removed. On average, the overall quality score decreases from 8.59 to 7.93 on the Mix dataset, and similar degradations are observed on the CS, Legal, and Agriculture datasets.

The most pronounced drops are consistently seen in the \textit{Comprehensiveness} and \textit{Empowerment} metrics. This is expected, as raw text chunks contain the detailed explanations, evidence, and nuanced language necessary for generating thorough and actionable answers. In contrast, a context composed solely of structured entity information, while semantically focused, lacks the narrative richness required by the LLM. These findings confirm our hypothesis: the hierarchical graph in LeanRAG acts as an effective semantic index and navigation system whose primary function is to precisely locate critical segments of unstructured text. The collaboration between structured graph traversal for guidance and the rich content of unstructured text for generation is essential to achieving state-of-the-art performance.

\begin{table}[t]
\scriptsize
\centering
\resizebox{0.45\textwidth}{!}{
\begin{tabular}{llcc}
\toprule
\textbf{Dataset} & \textbf{Metric} $\uparrow$ & \textbf{LeanRAG} & \textbf{LeanRAG w/o Context} \\
\midrule

\multirow{4}{*}{Mix}
& Comprehensiveness & \textbf{8.89$\pm$0.01} & 8.15$\pm$0.02 $\downarrow$ \\
& Empowerment       & \textbf{8.16$\pm$0.02} & 7.80$\pm$0.01 $\downarrow$ \\
& Diversity         & \textbf{7.73$\pm$0.01} & 7.26$\pm$0.02 $\downarrow$ \\
& Overall           & \textbf{8.59$\pm$0.01} & 7.93$\pm$0.01 $\downarrow$ \\
\midrule

\multirow{4}{*}{CS}
& Comprehensiveness & \textbf{8.92$\pm$0.01} & 8.66$\pm$0.02 $\downarrow$ \\
& Empowerment       & \textbf{8.68$\pm$0.02} & 8.19$\pm$0.03 $\downarrow$ \\
& Diversity         & \textbf{7.87$\pm$0.02} & 7.57$\pm$0.02 $\downarrow$ \\
& Overall           & \textbf{8.82$\pm$0.02} & 8.34$\pm$0.02 $\downarrow$ \\
\midrule

\multirow{4}{*}{Legal}
& Comprehensiveness & \textbf{8.88$\pm$0.02 } & 8.49$\pm$0.01 $\downarrow$ \\
& Empowerment       & \textbf{8.42$\pm$0.03} & 8.11$\pm$0.04 $\downarrow$ \\
& Diversity         & \textbf{7.49$\pm$0.03} & 7.09$\pm$0.04 $\downarrow$ \\
& Overall           & \textbf{8.49$\pm$0.04} & 8.00$\pm$0.04 $\downarrow$ \\
\midrule

\multirow{4}{*}{Agriculture}
& Comprehensiveness & \textbf{8.94$\pm$0.06} & 8.65$\pm$0.01 $\downarrow$ \\
& Empowerment       & \textbf{8.66$\pm$0.02} & 8.16$\pm$0.05 $\downarrow$ \\
& Diversity         & \textbf{8.06$\pm$0.03} & 7.88$\pm$0.05 $\downarrow$ \\
& Overall           & \textbf{8.87$\pm$0.02} & 8.53$\pm$0.03 $\downarrow$ \\
\bottomrule
\end{tabular}
}
\caption{Necessity analysis of textual context}
\label{tab:rq4_context_necessity}
\end{table}

\section{Conclusions}
To address the critical challenges of ``semantic islands'' and the structure-retrieval mismatch prevalent in the KG-based RAG systems, we introduced \textbf{LeanRAG}, a novel framework that resolves these issues through a tight co-design of its knowledge aggregation and retrieval mechanisms. Our approach features a hierarchical aggregation algorithm that constructs a fully navigable semantic network by generating explicit relations between abstract summary concepts, and a complementary bottom-up, LCA-based retrieval strategy that efficiently traverses this structure. Extensive experiments validated our design, demonstrating that LeanRAG achieves state-of-the-art performance while significantly reducing information redundancy. Furthermore, our ablation studies confirmed that both the generation of summary information and the original textual context are essential for producing comprehensive and diverse answers.

\section*{Acknowledgments}
The research was supported by Shanghai Artificial Intelligence Laboratory, the National Key R\&D Program of China (Grant No. 2022ZD0160201) and the Science and Technology Commission of Shanghai Municipality (Grant Nos. 22DZ1100102).

\bibliography{aaai2026}

\begin{thebibliography}{25}
\providecommand{\natexlab}[1]{#1}

\bibitem[{Chen et~al.(2024)Chen, Xiao, Zhang, Luo, Lian, and Liu}]{bge-m3}
Chen, J.; Xiao, S.; Zhang, P.; Luo, K.; Lian, D.; and Liu, Z. 2024.
\newblock BGE M3-Embedding: Multi-Lingual, Multi-Functionality, Multi-Granularity Text Embeddings Through Self-Knowledge Distillation.
\newblock arXiv:2402.03216.

\bibitem[{Edge et~al.(2024)Edge, Trinh, Cheng, Bradley, Chao, Mody, Truitt, Metropolitansky, Ness, and Larson}]{graphrag}
Edge, D.; Trinh, H.; Cheng, N.; Bradley, J.; Chao, A.; Mody, A.; Truitt, S.; Metropolitansky, D.; Ness, R.~O.; and Larson, J. 2024.
\newblock From local to global: A graph rag approach to query-focused summarization.
\newblock \emph{arXiv preprint arXiv:2404.16130}.

\bibitem[{Gao et~al.(2023)Gao, Xiong, Gao, Jia, Pan, Bi, Dai, Sun, Guo, Wang, and Wang}]{Gao2023RetrievalAugmentedGFA}
Gao, Y.; Xiong, Y.; Gao, X.; Jia, K.; Pan, J.; Bi, Y.; Dai, Y.; Sun, J.; Guo, Q.; Wang, M.; and Wang, H. 2023.
\newblock Retrieval-Augmented Generation for Large Language Models: A Survey.
\newblock \emph{ArXiv}, abs/2312.10997.

\bibitem[{Guo et~al.(2024)Guo, Xia, Yu, Ao, and Huang}]{lightrag}
Guo, Z.; Xia, L.; Yu, Y.; Ao, T.; and Huang, C. 2024.
\newblock LightRAG: Simple and Fast Retrieval-Augmented Generation.
\newblock \emph{arXiv preprint arXiv:2410.05779}.

\bibitem[{Huang et~al.(2025{\natexlab{a}})Huang, Huang, Yang, Pan, Chen, Ma, Chen, and Cheng}]{hirag}
Huang, H.; Huang, Y.; Yang, J.; Pan, Z.; Chen, Y.; Ma, K.; Chen, H.; and Cheng, J. 2025{\natexlab{a}}.
\newblock HiRAG: Retrieval-Augmented Generation with Hierarchical Knowledge.
\newblock arXiv:2503.10150.

\bibitem[{Huang et~al.(2025{\natexlab{b}})Huang, Yu, Ma, Zhong, Feng, Wang, Chen, Peng, Feng, Qin et~al.}]{huang2025survey}
Huang, L.; Yu, W.; Ma, W.; Zhong, W.; Feng, Z.; Wang, H.; Chen, Q.; Peng, W.; Feng, X.; Qin, B.; et~al. 2025{\natexlab{b}}.
\newblock A survey on hallucination in large language models: principles, taxonomy, challenges, and open questions.
\newblock \emph{ACM Transactions on Information Systems}, 43(2): 1--55.

\bibitem[{Izacard et~al.(2021)Izacard, Caron, Hosseini, Riedel, Bojanowski, Joulin, and Grave}]{izacard2021unsupervised}
Izacard, G.; Caron, M.; Hosseini, L.; Riedel, S.; Bojanowski, P.; Joulin, A.; and Grave, E. 2021.
\newblock Unsupervised dense information retrieval with contrastive learning.
\newblock \emph{arXiv preprint arXiv:2112.09118}.

\bibitem[{Jiang et~al.(2023)Jiang, Xu, Gao, Sun, Liu, Dwivedi-Yu, Yang, Callan, and Neubig}]{jiang2023active}
Jiang, Z.; Xu, F.~F.; Gao, L.; Sun, Z.; Liu, Q.; Dwivedi-Yu, J.; Yang, Y.; Callan, J.; and Neubig, G. 2023.
\newblock Active retrieval augmented generation.
\newblock In \emph{International Conference on Empirical Methods in Natural Language Processing (EMNLP)}, 7969--7992.

\bibitem[{Karpukhin et~al.(2020)Karpukhin, Oguz, Min, Lewis, Wu, Edunov, Chen, and Yih}]{karpukhin2020dense}
Karpukhin, V.; Oguz, B.; Min, S.; Lewis, P.~S.; Wu, L.; Edunov, S.; Chen, D.; and Yih, W.-t. 2020.
\newblock Dense Passage Retrieval for Open-Domain Question Answering.
\newblock In \emph{International Conference on Empirical Methods in Natural Language Processing (EMNLP)}, 6769--6781.

\bibitem[{Lewis et~al.(2020)Lewis, Perez, Piktus, Petroni, Karpukhin, Goyal, K{\"u}ttler, Lewis, Yih, Rockt{\"a}schel et~al.}]{naiverag}
Lewis, P.; Perez, E.; Piktus, A.; Petroni, F.; Karpukhin, V.; Goyal, N.; K{\"u}ttler, H.; Lewis, M.; Yih, W.-t.; Rockt{\"a}schel, T.; et~al. 2020.
\newblock Retrieval-augmented generation for knowledge-intensive nlp tasks.
\newblock \emph{Advances in neural information processing systems}, 33: 9459--9474.

\bibitem[{Li et~al.(2024)Li, Chen, Ren, Cheng, Zhao, yun Nie, and Wen}]{Li2024TheDAA}
Li, J.; Chen, J.; Ren, R.; Cheng, X.; Zhao, W.~X.; yun Nie, J.; and Wen, J.-R. 2024.
\newblock The Dawn After the Dark: An Empirical Study on Factuality Hallucination in Large Language Models.
\newblock In \emph{Annual Meeting of the Association for Computational Linguistics}, 10879--10899.

\bibitem[{Liang et~al.(2025)Liang, Bo, Gui, Zhu, Zhong, Zhao, Sun, Zhang, Zhou, Chen et~al.}]{kag}
Liang, L.; Bo, Z.; Gui, Z.; Zhu, Z.; Zhong, L.; Zhao, P.; Sun, M.; Zhang, Z.; Zhou, J.; Chen, W.; et~al. 2025.
\newblock Kag: Boosting llms in professional domains via knowledge augmented generation.
\newblock In \emph{Companion Proceedings of the ACM on Web Conference 2025}, 334--343.

\bibitem[{Liu et~al.(2024)Liu, Feng, Xue, Wang, Wu, Lu, Zhao, Deng, Zhang, Ruan et~al.}]{liu2024deepseek}
Liu, A.; Feng, B.; Xue, B.; Wang, B.; Wu, B.; Lu, C.; Zhao, C.; Deng, C.; Zhang, C.; Ruan, C.; et~al. 2024.
\newblock Deepseek-v3 technical report.
\newblock \emph{arXiv preprint arXiv:2412.19437}.

\bibitem[{Page et~al.(1999)Page, Brin, Motwani, and Winograd}]{pagerank}
Page, L.; Brin, S.; Motwani, R.; and Winograd, T. 1999.
\newblock The PageRank citation ranking: Bringing order to the web.
\newblock Technical report, Stanford infolab.

\bibitem[{Peng et~al.(2024)Peng, Zhu, Liu, Bo, Shi, Hong, Zhang, and Tang}]{Peng2024GraphRGA}
Peng, B.; Zhu, Y.; Liu, Y.; Bo, X.; Shi, H.; Hong, C.; Zhang, Y.; and Tang, S. 2024.
\newblock Graph Retrieval-Augmented Generation: A Survey.
\newblock \emph{ArXiv}, abs/2408.08921.

\bibitem[{Qian et~al.(2024)Qian, Zhang, Liu, Mao, and Dou}]{qian2024memorag}
Qian, H.; Zhang, P.; Liu, Z.; Mao, K.; and Dou, Z. 2024.
\newblock Memorag: Moving towards next-gen rag via memory-inspired knowledge discovery.
\newblock \emph{arXiv preprint arXiv:2409.05591}, 1.

\bibitem[{Reynolds(2015)}]{reynolds2015gaussian}
Reynolds, D. 2015.
\newblock Gaussian mixture models.
\newblock In \emph{Encyclopedia of biometrics}, 827--832. Springer.

\bibitem[{Robertson, Zaragoza et~al.(2009)}]{robertson2009probabilistic}
Robertson, S.; Zaragoza, H.; et~al. 2009.
\newblock The probabilistic relevance framework: BM25 and beyond.
\newblock \emph{Foundations and Trends{\textregistered} in Information Retrieval}, 3(4): 333--389.

\bibitem[{Sarthi et~al.(2024)Sarthi, Abdullah, Tuli, Khanna, Goldie, and Manning}]{sarthi2024raptor}
Sarthi, P.; Abdullah, S.; Tuli, A.; Khanna, S.; Goldie, A.; and Manning, C.~D. 2024.
\newblock Raptor: Recursive abstractive processing for tree-organized retrieval.
\newblock In \emph{International Conference on Learning Representations (ICLR)}.

\bibitem[{Tonellotto et~al.(2024)Tonellotto, Trappolini, Silvestri, Campagnano, Siciliano, Cuconasu, Maarek, and Filice}]{Tonellotto2024ThePOA}
Tonellotto, N.; Trappolini, G.; Silvestri, F.; Campagnano, C.; Siciliano, F.; Cuconasu, F.; Maarek, Y.; and Filice, S. 2024.
\newblock The Power of Noise: Redefining Retrieval for RAG Systems.
\newblock \emph{ACM International Conference on Research and Development in Information Retrieval (SIGIR)}.

\bibitem[{Wang et~al.(2024)Wang, Wang, Gao, Zhang, Wu, Xu, Shi, Wang, Li, Qian et~al.}]{wang2024searching}
Wang, X.; Wang, Z.; Gao, X.; Zhang, F.; Wu, Y.; Xu, Z.; Shi, T.; Wang, Z.; Li, S.; Qian, Q.; et~al. 2024.
\newblock Searching for best practices in retrieval-augmented generation.
\newblock \emph{arXiv preprint arXiv:2407.01219}.

\bibitem[{Wang et~al.(2025)Wang, Wang, Le, Zheng, Mishra, Perot, Zhang, Mattapalli, Taly, Shang, Lee, and Pfister}]{ICLR2025_2ea06b52}
Wang, Z.~R.; Wang, Z.; Le, L.; Zheng, H.~S.; Mishra, S.; Perot, V.; Zhang, Y.; Mattapalli, A.; Taly, A.; Shang, J.; Lee, C.-Y.; and Pfister, T. 2025.
\newblock Speculative RAG: Enhancing Retrieval Augmented Generation through Drafting.
\newblock In Yue, Y.; Garg, A.; Peng, N.; Sha, F.; and Yu, R., eds., \emph{International Conference on Representation Learning}, volume 2025, 18483--18505.

\bibitem[{Xu et~al.(2024)Xu, Cruz, Guevara, Wang, Deshpande, Wang, and Li}]{Xu2024RetrievalAugmentedGWA}
Xu, Z.; Cruz, M.~J.; Guevara, M.; Wang, T.; Deshpande, M.; Wang, X.; and Li, Z. 2024.
\newblock Retrieval-Augmented Generation with Knowledge Graphs for Customer Service Question Answering.
\newblock \emph{ACM International Conference on Research and Development in Information Retrieval (SIGIR)}.

\bibitem[{Yang et~al.(2025)Yang, Li, Yang, Zhang, Hui, Zheng, Yu, Gao, Huang, Lv et~al.}]{yang2025qwen3technicalreport}
Yang, A.; Li, A.; Yang, B.; Zhang, B.; Hui, B.; Zheng, B.; Yu, B.; Gao, C.; Huang, C.; Lv, C.; et~al. 2025.
\newblock Qwen3 Technical Report.
\newblock arXiv:2505.09388.

\bibitem[{Zhao et~al.(2024)Zhao, Zhang, Yu, Wang, Geng, Fu, Yang, Zhang, and Cui}]{Zhao2024RetrievalAugmentedGFA}
Zhao, P.; Zhang, H.; Yu, Q.; Wang, Z.; Geng, Y.; Fu, F.; Yang, L.; Zhang, W.; and Cui, B. 2024.
\newblock Retrieval-Augmented Generation for AI-Generated Content: A Survey.
\newblock \emph{ArXiv}, abs/2402.19473.

\end{thebibliography}


\appendix
\section*{Appendix}
The appendix provides supplementary materials and detailed information to support the main findings of this paper. It includes a comprehensive breakdown of our methodology, covering the following key sections: More Results of Model Performance Across Four Datasets, Experimental Implementation Details, QA Cases of LeanRAG, Prompt Templates used in LeanRAG.

This section is designed to ensure the reproducibility of our work by offering an in-depth look at the implementation specifics, including a detailed description of the datasets, the graph construction process, and the retrieval strategies employed. The provided prompt examples and templates further illustrate the mechanisms used to guide the Large Language Model (LLM) in generating high-quality responses.

\section{A. More Results of Model Performance Across Four Datasets}
The results consistently demonstrate that LeanRAG outperforms all baseline methods across all four datasets and evaluation metrics, achieving notably higher win rates. This suggests that the proposed approach offers a more efficient and reliable framework for retrieval and generation compared to existing methods.

LeanRAG significantly outperforms NaiveRAG, FastGraphRAG, and KAG: Across these baselines, LeanRAG exhibits overwhelming superiority, with win rates often exceeding 95\%, and reaching 100\% in some cases. This advantage is particularly pronounced in the Empowerment and Diversity metrics, underscoring LeanRAG’s ability to leverage structured knowledge graphs to provide more relevant and diverse information. These findings validate the fundamental advantage of graph-based methods over simple text retrieval approaches.

LeanRAG demonstrates strong performance against more advanced baselines such as GraphRAG, LightRAG, and HiRAG: Although the win rates are comparatively lower than against simpler baselines, LeanRAG still maintains a substantial performance margin. When compared to other graph-based methods like GraphRAG and HiRAG, LeanRAG achieves win rates consistently between 50\% and 80\%. This highlights the competitive advantage of LeanRAG’s strategy in aggregating entities and constructing multi-level semantic networks, surpassing conventional graph-based or hierarchy-based RAG techniques. On the Comprehensiveness metric within the Legal domain, LeanRAG’s win rate against GraphRAG is the lowest (51.0\% vs. 49.0\%), indicating that the dense and domain-specific nature of legal texts poses similar challenges for both models. In comparison with LightRAG, LeanRAG consistently outperforms it across the Mix and Legal datasets, with win rates frequently exceeding 80\%, particularly in the Empowerment and Diversity categories. This indicates that LeanRAG’s enhanced graph construction and retrieval mechanisms are more effective than those employed by LightRAG.

Consistency across datasets and metrics: LeanRAG’s superior performance is not limited to any single dataset or metric. It consistently outperforms baselines across diverse domains, including Mix, Computer Science, Legal, and Agriculture, demonstrating both robustness and generalizability. Notably, its highest win rates are often observed in the Empowerment and Diversity metrics, which are critical for generating high-quality, non-redundant, and actionable responses. This underscores the effectiveness of LeanRAG’s core design in producing meaningful outputs.

\begin{table*}[htb]
\scriptsize
\caption{Win rates (\%) of LeanRAG, its two variants (for ablation study), and baseline methods on QFS tasks.}

\resizebox{\textwidth}{!}{
\begin{tabular}{@{}lcccccccc@{}}
\toprule
\textbf{}    & \multicolumn{2}{c}{\textbf{Mix}} & \multicolumn{2}{c}{\textbf{CS}} & \multicolumn{2}{c}{\textbf{Legal}} & \multicolumn{2}{c}{\textbf{Agriculture}} \\
\midrule
                & NaiveRAG & \textbf{LeanRAG}& NaiveRAG & \textbf{LeanRAG} & NaiveRAG & \textbf{LeanRAG}& NaiveRAG & \textbf{LeanRAG} \\
\cmidrule(lr){2-3}  \cmidrule(lr){4-5} \cmidrule(lr){6-7} \cmidrule(lr){8-9}
Comprehensiveness& 11.9\%& \underline{88.1\%}& 41.0\%& \underline{59.0\%}& 30.0\%& \underline{70.0\%}& 34.0\%& \underline{66.0\%}\\
Empowerment& 1.5\%& \underline{98.5\%}& 40.5\%& \underline{59.5\%}& 24.5\%& \underline{75.5\%}& 15.5\%&\underline{84.5\%}\\
Diversity& 3.1\%& \underline{96.9\%}& 28.0\%& \underline{72.0\%}& 9.0\%& \underline{91.0\%}& 10.0\%&\underline{90.0\%}\\
Overall& 2.7\%& \underline{97.3\%}& 39.5\%& \underline{60.5\%}& 23.5\%& \underline{76.5\%}& 16.0\%&\underline{84.0\%}\\
\midrule
                & GraphRAG & \textbf{LeanRAG}& GraphRAG & \textbf{LeanRAG} & GraphRAG & \textbf{LeanRAG}& GraphRAG & \textbf{LeanRAG}\\
\cmidrule(lr){2-3}  \cmidrule(lr){4-5} \cmidrule(lr){6-7} \cmidrule(lr){8-9}
Comprehensiveness& 35.0\%& \underline{65.0\%}& 41.0\%& \underline{59.0\%}& 49.0\%& \underline{51.0\%}& 45.5\%&\underline{54.5\%}\\
Empowerment& 20.0\%& \underline{80.0\%}& 33.5\%& \underline{66.5\%}& 44.0\%& \underline{56.0\%}& 27.0\%&\underline{73.0\%}\\
Diversity& 16.5\%& \underline{83.5\%}& 34.0\%& \underline{66.0\%}& 44.0\%& \underline{56.0\%}& 22.0\%&\underline{78.0\%}\\
Overall& 21.9\%& \underline{78.1\%}& 37.5\%& \underline{62.5\%}& 47.0\%& \underline{53.0\%}& 28.5\%&\underline{71.5\%}\\
\midrule
                & LightRAG & \textbf{LeanRAG}& LightRAG & \textbf{LeanRAG} & LightRAG & \textbf{LeanRAG}& LightRAG & \textbf{LeanRAG}\\
\cmidrule(lr){2-3}  \cmidrule(lr){4-5} \cmidrule(lr){6-7} \cmidrule(lr){8-9}
Comprehensiveness& 28.8\%& \underline{71.2\%}& 44.5\%& \underline{55.5\%}& 25.0\%& \underline{75.0\%}& 38.0\%&\underline{62.0\%}\\
Empowerment& 16.5\%& \underline{83.5\%}& 35.5\%& \underline{64.5\%}& 12.0\%& \underline{88.0\%}& 17.0\%&\underline{83.0\%}\\
Diversity& 13.1\%& \underline{86.9\%}& 34.0\%& \underline{66.0\%}& 40.5\%&\underline{59.5\%} & 16.5\%&\underline{83.5\%}\\
Overall& 18.8\%& \underline{81.2\%}& 38.5\%& \underline{61.5\%}& 21.0\%& \underline{79.0\%}& 18.5\%&\underline{81.5\%}\\
\midrule
                & FastGraphRAG & \textbf{LeanRAG}& FastGraphRAG & \textbf{LeanRAG} & FastGraphRAG & \textbf{LeanRAG}& FastGraphRAG & \textbf{LeanRAG}\\
\cmidrule(lr){2-3}  \cmidrule(lr){4-5} \cmidrule(lr){6-7} \cmidrule(lr){8-9}
Comprehensiveness& 0\%& \underline{100\%}& 0.5\%& \underline{99.5\%}& 1.0\%& \underline{99.0\%}& 0.5\%&\underline{99.5\%}\\
Empowerment& 0\%& \underline{100\%}& 0.0\%& \underline{100.0\%}& 0.5\%& \underline{99.5\%}& 0.0\%&\underline{100.0\%}\\
Diversity& 0\%& \underline{100\%}& 0.8\%& \underline{99.2\%}& 2.5\%& \underline{97.5\%}& 0.0\%&\underline{100.0\%}\\
Overall& 0\%& \underline{100\%}& 0.0\%& \underline{100.0\%}& 4.5\%& \underline{95.5\%}& 0.0\%&\underline{100.0\%}\\
\midrule
                & KAG & \textbf{LeanRAG}& KAG & \textbf{LeanRAG} & KAG & \textbf{LeanRAG}& KAG & \textbf{LeanRAG}\\
\cmidrule(lr){2-3}  \cmidrule(lr){4-5} \cmidrule(lr){6-7} \cmidrule(lr){8-9}
Comprehensiveness& 1.5\%& \underline{98.5\%}& 5.0\%& \underline{95.0\%}& 5.0\%& \underline{95.0\%}& 2.5\%&\underline{97.5\%}\\
Empowerment& 1.9\%& \underline{98.1\%}& 3.0\%& \underline{97.0\%}& 4.5\%& \underline{95.5\%}& 2.5\%&\underline{97.5\%}\\
Diversity& 1.2\%& \underline{98.8\%}& 4.0\%& \underline{96.0\%}& 2.5\%& \underline{97.5\%}& 1.0\%&\underline{99.0\%}\\
Overall& 1.2\%& \underline{98.8\%}& 3.5\%& \underline{96.5\%}& 4.5\%& \underline{95.5\%}& 1.0\%&\underline{99.0\%}\\
\midrule
                & HiRAG & \textbf{LeanRAG}& HiRAG & \textbf{LeanRAG} & HiRAG & \textbf{LeanRAG}& HiRAG & \textbf{LeanRAG}\\
\cmidrule(lr){2-3}  \cmidrule(lr){4-5} \cmidrule(lr){6-7} \cmidrule(lr){8-9}
Comprehensiveness& 43.8\%& \underline{56.2\%}& 46.5\%& \underline{53.5\%}& 29.5\%& \underline{70.5\%}& 49.5\%&\underline{50.5\%}\\
Empowerment& 26.5\%& \underline{73.5\%}& 43.5\%& \underline{56.5\%}& 16.5\%& \underline{83.5\%}& 26.5\%&\underline{73.5\%}\\
Diversity& 20.4\%& \underline{79.6\%}& 44.5\%& \underline{55.5\%}& 23.5\%& \underline{76.5\%}& 23.5\%&\underline{76.5\%}\\
Overall& 28.1\%& \underline{71.9\%}& 45.0\%& \underline{55.0\%}& 21.5\%& \underline{78.5\%}& 28.0\%&\underline{72\%}\\
\bottomrule
\end{tabular}
\label{tab:rq1winrate}
}
\end{table*}

\section{B. Experimental Implementation Details}
\subsection{B.1 Dataset Details}
This subsection provides a comprehensive description of the dataset(s) utilized in this study. It includes details regarding the source of the data and its overall size (e.g., number of documents, total tokens).
\begin{table}[!htbp]
	\caption{Statistics of task datasets.}
	\setlength\tabcolsep{6pt}
	\renewcommand{\arraystretch}{1.2}
	\centering
	\resizebox{0.48\textwidth}{!}{
		\begin{tabular}{lcccl}
			\hline
			\textbf{Dataset}& \textbf{Mix}& \textbf{CS}&\textbf{Legal}&\textbf{Agriculture}\\
			\hline
			 \# of Documents& 61& 10&94&12\\
			\# of Tokens& 625,948&2,210,894&5,279,400&2,028,496\\
                \hline
		\end{tabular}
		\label{tab:datasets_statistics}
	}
\end{table}

As presented in Table \ref{tab:datasets_statistics}, the datasets vary significantly in size and content. The Legal dataset is the largest, containing 94 documents and a substantial 5,279,400 tokens, reflecting the detailed and extensive nature of legal texts. In contrast, the CS (Computer Science) dataset, while having fewer documents (10), still comprises a significant 2,210,894 tokens, indicating potentially longer and more technical documents within that domain. The Agriculture dataset contributes 12 documents and 2,028,496 tokens, while the Mix dataset, serving as a general collection, includes 61 documents and 625,948 tokens. These diverse characteristics allow for a thorough assessment of our model's performance across varied information landscapes.

\subsection{B.2 Graph Construction Implementation Details}

To effectively manage the scale of LeanRAG, we introduced a hyperparameter, $clustersize$, which allows us to control the number of clusters generated during the Gaussian Mixture Model (GMM) clustering process by manually limiting the number of nodes within each cluster. This design choice provides a significant degree of controllability, enabling us to adjust the size of LeanRAG according to specific application requirements.

In our experiments, we performed a unified entity and relationship extraction for all documents within each dataset to build a single knowledge graph. This approach ensures a consistent graph structure for each dataset, rather than generating a separate graph for each question-answer pair.

Despite the four datasets varying considerably in both size and domain, we consistently used a $clustersize$ of 20 for graph construction. It's important to note that $clustersize$ is a pivotal factor that not only dictates the overall size of the LeanRAG graph but also profoundly impacts its retrieval efficiency and quality. Beyond this, the threshold $\tau$, which governs the generation of inter-cluster relationships, also profoundly impacts LeanRAG's performance. For our experiments, we set this threshold to 3.To further assess the efficacy of our method and cater to diverse use cases, future work will involve a comprehensive exploration of how different $clustersize$ values and $\tau$ values influence LeanRAG's performance.

\subsection{B.3 Graph Retrieval Details}
\subsubsection{B.3.1 Chunk Selection Strategy}
Based on our observations of traditional GraphRAG methods, we found that even after extracting structured entities, relationships, and community information, the original text chunks remain crucial for answering questions. This is because these chunks often contain incoherent semantic information that cannot be structurally extracted, yet still plays a vital role. Consequently, in LeanRAG, we also return the top-C retrieved chunks during the process.

Our specific approach is as follows: After identifying the initial seed nodes  $V_{\text{seed}}$, we trace back to their original text chunks. We then rank these chunks in descending order based on the number of entities from  $V_{\text{seed}}$ that appear within each chunk. Finally, we return the top-C chunks from this ranked list. This method allows us to pinpoint the top-C chunks most relevant to the query by aligning with the user's intent through entity-based searching, which we find to be more effective than the similarity-based chunk retrieval employed by Naive RAG.

\subsection{B.4 Experiment Settings}
To ensure our method achieves optimal performance across all four datasets, we fine-tuned the hyperparameter $clustersize$, $Top-N$, and $Top-C$. The specific parameter settings used for these adjustments are detailed below:

\begin{table}[thb]
    \small
	\caption{Setting of task datasets.}
	\setlength\tabcolsep{6pt}
	\renewcommand{\arraystretch}{1.2}
	\centering
	\resizebox{0.48\textwidth}{!}{
		\begin{tabular}{ccccc}
			\hline
			\textbf{Dataset}& \textbf{Mix}& \textbf{CS}&\textbf{Legal}&\textbf{Agriculture}\\
			\hline
			 $clustersize$& 20& 20&20&20\\
			 $N$& 10&10&15&10\\
             $C$& 5&10&10&5\\
                \hline
		\end{tabular}
		\label{tab:datasets_seeting}
	}
\end{table}

Our observations during the retrieval phase revealed distinct characteristics across the datasets, influencing the optimal parameter settings for effective information retrieval.

Specifically, for the Mix and Agriculture datasets, a relatively smaller number of seed nodes $V_{\text{seed}}$was sufficient for robust query resolution. This can be attributed to the limited scope of content within a subset of documents and the overall stronger internal connectedness within their respective knowledge bases.

Conversely, the Computer Science (CS) dataset presented unique challenges. Its weaker intrinsic associativity and the less structured nature of its specialized terminology necessitated the retrieval of a larger number of supporting chunks. This suggests that relevant information for a given query in the CS domain might be more distributed and less directly interlinked within the graph structure.

Finally, the Legal dataset, characterized by its highly specialized and extensive terminology and greater document-level separability, required the retrieval of a larger volume of information. This indicated a need for a higher count of $V_{\text{seed}}$to achieve a comprehensive understanding of the query, as pertinent details tended to be more dispersed across a broader range of documents.

\section{C. QA Cases of LeanRAG}

To illustrate the effectiveness of our approach, this section presents a few straightforward examples comparing the performance of LeanRAG with the HiRAG method. These cases are designed to highlight how LeanRAG's optimized graph structure and retrieval strategy lead to more precise and coherent answers. By directly contrasting their outputs, we aim to demonstrate the practical benefits of our method in various query scenarios,the case can be found in Table \ref{tab:example_assessment}.

\begin{table*}[h]
    \caption{Example question, answers, and LLM-generated assessments for the Mix dataset.}
    \small
    \centering
    \def\arraystretch{1.25}%
    \begin{tabular}{>{\raggedright\arraybackslash}p{0.08\textwidth}>{\raggedright\arraybackslash}p{0.92\textwidth}} \hline 
 \textbf{Question} & \textbf{What is the significance of studying puns in the context of human-computer interaction and machine translation?}\\ \hline 
 \textbf{LeanRAG} & The study of puns in the context of human-computer interaction (HCI) and machine translation (MT) holds significant importance for several reasons, as evidenced by the provided data tables and related research.

\textbf{1. Enhancing Natural Language Understanding} [...] \textbf{2. Improving Machine Translation Systems} [...]
 \textbf{3. Benchmarking and Model Evaluation} [...] \textbf{4. Cross-Disciplinary Applications} [...] \textbf{5. Challenges and Future Directions} [...]

In summary, studying puns contributes to deeper language understanding, better translation systems, and more sophisticated human-computer interactions by addressing ambiguity, humor, and cultural context—key hurdles in NLP.\vspace*{2mm}\\ \hline 
\textbf{HiRAG} & Significance of Studying Puns in Human-Computer Interaction and Machine Translation

\textbf{Human-Computer Interaction} [...] \textbf{Machine Translation} [...] \textbf{Real-World Applications} [...] 
In summary, the study of puns in the context of human-computer interaction and machine translation is significant for developing more sophisticated and effective NLP systems. By enabling machines to understand and handle puns, we can create more natural and engaging interactions and more accurate translations, thereby enhancing the overall utility and effectiveness of these technologies.
\vspace*{2mm}\\ \hline 
\textbf{LLM \break Decision} & 
\textbf{Comprehensiveness: Winner=1 (LeanRAG)}. Answer 1 provides a more detailed and structured analysis of the significance of studying puns in HCI and machine translation. It breaks down the topic into specific areas such as natural language understanding, machine translation systems, benchmarking, cross-disciplinary applications, and future challenges. This level of detail makes it more comprehensive compared to Answer 2, which, while informative, lacks the depth and structured approach of Answer 1.
\vspace*{2mm}
\newline
\textbf{Empowerment: Winner=1 (LeanRAG)}. Answer 1 empowers the reader by not only explaining the significance of studying puns but also by discussing specific tools (e.g., CMU Pronouncing Dictionary), models (e.g., Bidirectional RNN, Hidden Markov Model), and datasets (e.g., HOMOGRAPHIC DATASET, HETEROGRAPHIC DATASET) used in the research. This provides the reader with concrete information to further explore the topic or apply the knowledge in practical scenarios. Answer 2, while informative, does not provide as much actionable or specific information.
\vspace*{2mm}
\newline
\textbf{Diversity: Winner=1 (LeanRAG)}. Answer 1 offers a richer diversity of perspectives and insights by covering multiple aspects such as natural language understanding, machine translation, benchmarking, cross-disciplinary applications, and future challenges. It also mentions specific models and datasets, adding layers of depth and variety to the discussion. Answer 2, while covering the basics well, does not provide the same level of varied insights or detailed examples.
\vspace*{2mm}
\newline
\textbf{Overall Winner: Winner=1 (LeanRAG)}. Answer 1 is the overall winner because it excels in comprehensiveness, empowerment, and diversity. It provides a detailed, structured, and multi-faceted analysis of the topic, equips the reader with specific tools and models for further exploration, and offers a wide range of perspectives and insights. Answer 2 is informative but lacks the depth, specificity, and variety that make Answer 1 superior.\vspace*{2mm}\\ \hline
    \end{tabular}
    \label{tab:example_assessment}
\end{table*}

\section{D. Prompt Templates used in LeanRAG}
This section details the specific prompt templates employed within the LeanRAG framework. While our knowledge graph (KG) generation code aligns with that of LightRAG and will not be reiterated here, this chapter focuses on the four distinct prompt templates critical to LeanRAG's operation: the Entity Aggregation Prompt, the Inter-Cluster Relation Generation Prompt, the Score Scoring Prompt, and the Win Rate Evaluation Prompt. Each prompt plays a vital role in guiding the Large Language Model (LLM) through various stages of information processing, from consolidating entities to evaluating retrieval outcomes.

\subsection{D.1 Prompt Templates for Entity Aggregation}
As depicted in Figure \ref{tab:entity_aggregation_prompt}, we leverage the clusters generated by the Gaussian Mixture Model (GMM) to derive descriptions of all entities within a cluster, along with the relationships between these intra-cluster entities. This information is then used to generate an aggregated entity. To circumvent the limitations of traditional community concepts, which can forcibly aggregate all entities and inadvertently assign irrelevant attributes, we explicitly constrain the Large Language Model (LLM) to generate information solely based on the current set of entity descriptions. Furthermore, we emphasize the connecting role of the generated aggregated entity for its constituent sub-entities, ensuring its relevance and coherence within the broader knowledge graph.

\begin{table*}[htbp]
\small
\centering
\caption{The prompt template of aggregate entities from entity cluster.}
\label{tab:entity_aggregation_prompt}
\begin{tabular}{p{0.95\linewidth}}
\toprule
\textbf{Entity aggregation prompt} \\ 
\midrule

\begin{minipage}[t]{\linewidth}
\textbf{Role: Entity Aggregation Analyst}

\vspace{1em}

\textbf{Profile} \\
- author: LangGPT   \\
- version: 1.1  \\
- language: English \\
- description: You are an expert in \textbf{concept synthesis} and entity aggregation. Your task is to identify a meaningful aggregate entity from a set of related entities and extract structured, comprehensive insights based solely on provided evidence.

\vspace{1em}

\textbf{Skills} \\
- Abstraction and naming of collective concepts based on entity roles, and relationships    \\
- Structured summarization and typology recognition \\
- Comparative and relational analysis across multiple entities  \\
- Strict grounding to provided data (no hallucinated content)   \\
- Extraction of both explicit and implicit shared characteristics     \\

\vspace{1em}

\textbf{Goals}

- Derive a meaningful aggregate entity that broadly represents the given entity set, capturing both explicit and nuanced connections    \\
- The aggregate entity name must not match any single entity in the set \\
- Provide an accurate, comprehensive, and concise description of the aggregate entity reflecting shared characteristics, structure, functions, and significance \\
- Extract as many structured findings as possible (at least 5, but preferably more) about the entity set based on grounded evidence, including roles, relationships, patterns, and unique features  \\

\vspace{1em}

\textbf{Output Format}  \\
- All output MUST be in a well-formed JSON-formatted string, strictly following the structure below.    \\
- Do NOT include any explanation, markdown, or extra text outside the JSON. \\

\vspace{0.5em}

Format: \\
Input: \texttt{\{input\_text\}} \\ 

Output:

\begin{verbatim}
{
  "entity_name": "<name>",
  "entity_description": "<description summarizing the shared traits, structure,
                         functions, and significance of the aggregation>",
  "findings": [
    {
      "summary": "<summary>",
      "explanation": "<explanation>"
    }
  ]
}

\end{verbatim}

\vspace{0.5em}

\textbf{Rules}

- Grounding Rule: All content must be based solely on the provided entity set — no external assumptions \\
- Naming Rule: The aggregate entity name must not be identical to any single entity; it should reflect a composite structure, function, or theme    \\
- Each finding must include a concise summary and a detailed explanation    \\
- Include findings about entity roles, interconnections, patterns, and any notable diversity or specialization within the set   \\
- Avoid adding speculative or unsupported interpretations

\vspace{1em}

\textbf{Workflows}  \\
1. Review the list of entities, focusing on types, descriptions, and relational structure   \\
2. Synthesize a generalized name that best represents the full entity set, emphasizing collective identity and function \\
3. Write a clear, evidence-based, and information-rich description of the aggregate entity  \\
4. Extract and elaborate on key findings, emphasizing structure, purpose, interconnections, diversity, and any emergent properties, and explicitly relate these to the contributions of the sub-entities

\end{minipage} \\

\bottomrule
\end{tabular}
\end{table*}

\subsection{D.2 Prompt Templates for Relation Aggregation}
As illustrated in Figure \ref{tab:relation_aggregation_prompt}, we employ a specialized relation prompt to generate relationships between the aggregated entities. This prompt leverages the names and descriptions of two aggregated entities, alongside the existing relationships between their constituent sub-entities, to infer and generate all relevant connections between the two aggregated entities. Given that the descriptions of the aggregated entities already encapsulate the broad information of their sub-entities, we did not incorporate additional sub-entity descriptions to enrich the input. Through this generation of relationships between sets of aggregated entities, LeanRAG effectively mitigates the problem of ``semantic islands'', thereby constructing a multi-level navigable semantic network.

\begin{table*}[htbp]
\small
\centering
\caption{The prompt template of generate relation between aggregation entities.}
\label{tab:relation_aggregation_prompt}
\begin{tabular}{p{0.95\linewidth}}
\toprule
\textbf{Relation aggregation prompt} \\ 
\midrule

\begin{minipage}[t]{\linewidth}
\textbf{Role: Inter-Aggregation Relationship Analyst}

\vspace{1em}

\textbf{Profile}
- author: LangGPT   \\
- version: 1.2      \\
- language: English \\
- description: You specialize in analyzing relationships between two aggregation entities. Your goal is to synthesize a high-level, abstract summary sentence that comprehensively covers all types of relationships between the sub-entities of two named aggregations, based solely on their descriptions and sub-entity relationships.

\vspace{1em}

\textbf{Skills}
- Aggregated reasoning across entity groups   \\
- Abstraction and synthesis of all cross-entity relationship types  \\
- Formal summarization under strict constraints \\
- Strong grounding without repetition or speculation    \\

\vspace{1em}

\textbf{Goals}
- Produce a summary ($\leq$\ {tokens\} words) that comprehensively and collectively covers all types of relationships between the sub-entities of Aggregation A and Aggregation B   \\
- Ensure the summary reflects the full diversity and scope of the sub-entity relationships, not just a single aspect    \\
- Avoid reproducing individual sub-entity relationships \\
- Emphasize structural, functional, or thematic connections at the group level

\vspace{1em}

\textbf{Input Format}

Aggregation A Name: \{entity\_a\} \\
Aggregation A Description: \{entity\_a\_description\}   \\
Aggregation B Name: \{entity\_b\}   \\
Aggregation B Description: \{entity\_b\_description\}   \\
Sub-Entity Relationships: \{relation\_information\}     \\

\vspace{1em}

\textbf{Output Format}

\textless Single-sentence explanation ($\leq$ {tokens} words) summarizing the relationship between Aggregation A and Aggregation B. Use abstract group-level language. The sentence must comprehensively reflect all types of relationships present between the sub-entities.\textgreater \\

\textbf{Rules}

- DO NOT name specific sub-entities (e.g., individuals) \\
- DO NOT use the term “community”; always refer to “aggregation,” “group,” “collection,” or thematic equivalents    \\
- DO use collective terms (e.g., “external reviewers,” “trade policy actors”)   \\
- The sentence must be $\leq$ \{tokens\} words, factual, grounded, and in formal English    \\
- The relationship must reflect an **aggregation-level abstraction**, such as:  \\
    - support/collaboration \\
    - review/feedback   \\
    - functional alignment  \\
    - domain linkage (e.g., one produces work, the other evaluates it)  \\
    - any other relevant relationship types present in the sub-entity relationships \\
- The summary must comprehensively cover the diversity and scope of all sub-entity relationships, not just a single type    \\

\vspace{1em}

\textbf{Example} \\
Input:  \\
Aggregation A Name: WTO External Contributors   \\
Aggregation A Description: A group of economists and trade policy experts who provided feedback on early drafts of WTO reports.  \\
Aggregation B Name: WTO Flagship Reports   \\
Aggregation B Description: Core analytical publications from the WTO addressing international trade issues.    \\

Sub-Entity Relationships:   \\
- External contributors provided expert review and feedback on preliminary drafts of flagship reports.  \\
- Feedback from the group was incorporated to enhance report quality and analytical depth.  \\

\vspace{1em}
Output: \\
The WTO External Contributors aggregation enhanced the analytical rigor and credibility of the WTO Flagship Reports aggregation by providing expert review, feedback, and collaborative input across multiple report drafts.
}
\end{minipage} \\

\bottomrule 
\end{tabular}
\end{table*}

\subsection{D.3 Prompt Template for Absolute Quality Scoring}

To obtain a quantitative measure of performance for each model, we designed a prompt for absolute quality scoring. This prompt instructs an evaluating LLM to assess a single generated answer based on our predefined metrics (\textit{Comprehensiveness}, \textit{Empowerment}, etc.) and assign a numerical score from 1 to 10 for each. To ensure transparency and facilitate analysis, the LLM is also required to provide a concise rationale for each score. All assessments are structured in a JSON format to ensure consistency and ease of parsing. The detailed template used for this scoring task is presented in Table~\ref{tab:score_prompt}.

\begin{table*}[htbp]
\small
\centering
\caption{The prompt template of scoring model response.}
\label{tab:score_prompt}
\begin{tabular}{p{0.95\linewidth}}
\toprule
\textbf{QA scoring prompt} \\ 
\midrule
\begin{minipage}[t]{\linewidth}

Your task is to evaluate the following answer based on four criteria. For each criterion, assign a score from 1 to 10 , following the detailed scoring rubric.  \\

\vspace{1em}

When explaining your score, you must refer directly to specific parts of the answer to justify your reasoning. Avoid general statements — your explanation must be grounded in the content provided.    \\

\vspace{1em}

- \textbf{Comprehensiveness}:   \\
How much detail does the answer provide to cover all aspects and details of the question?   \\

\vspace{1em}

- \textbf{Diversity}:   \\
How varied and rich is the answer in providing different perspectives and insights on the question? \\

\vspace{1em}

- \textbf{Empowerment}: \\
How well does the answer help the reader understand and make informed judgments about the topic?    \\

\vspace{1em}

- \textbf{Overall Quality}: \\
Provide an overall evaluation based on the combined performance across all four dimensions. Consider both content quality and answer usefulness to the question.    \\

\vspace{1em}

\textbf{Scoring Guidelines}:

\vspace{1em}

``1-2'': ``Low score description: Clearly deficient in this aspect, with significant issues.'',     \\
``3-4'': ``Below average score description: Lacking in several important areas, with noticeable problems.'',    \\
``5-6'': ``Average score description: Adequate but not exemplary, meets basic expectations with some minor issues.'',   \\
``7-8'': ``Above average score description: Generally strong but with minor shortcomings.'', \\
``9-10'': ``High score description: Outstanding in this aspect, with no noticeable issues.''   \\

\vspace{1em}

Here is the question:   \\
\{\texttt{query}\}  \\

\vspace{1em}

Here are the  answer:   \\

\{\texttt{answer}\}

\vspace{1em}

Evaluate the answer using the  criteria listed above and provide detailed explanations for each criterion with reference to the text. 
Output your evaluation in the following JSON format:

\begin{verbatim}
{{
    ``Comprehensiveness'': {{
        ``score'': ``[1-10]'',
        ``Explanation'': ``[Provide explanation here]''
    }},
    ``Empowerment'': {{
        ``score'': ``[1-10]'',
        ``Explanation'': ``[Provide explanation here]''
    }},
    ``Diversity'': {{
        ``score'': ``[1-10]'',
        ``Explanation'': ``[Provide explanation here]''
    }},
    ``Overall Quality'': {{
        ``score'': ``[1-10]'',
        ``Explanation'': ``[Summarize why this answer is the overall winner 
                        based on the three criteria]''
    }}
}}
\end{verbatim}

\end{minipage} \\

\bottomrule 
\end{tabular}
\end{table*}

\subsection{D.4 Prompt Template for Pairwise Comparison}

In addition to absolute scoring, we conducted pairwise comparisons to determine the relative performance between different models, resulting in win-rate statistics. For this purpose, we developed a separate prompt that presents the answers from two different models (e.g., LeanRAG vs. HiRAG) to an evaluating LLM. The prompt then instructs the evaluator to act as an impartial judge and determine which of the two answers is superior, considering the overall quality. The LLM must declare a ``winner'' and provide a detailed justification for its decision, again in a structured JSON format. The template used for these head-to-head comparisons is shown in Table~\ref{tab:rate_prompt}.

\begin{table*}[htbp]
\small
\centering
\caption{The prompt template of rating model response.}
\label{tab:rate_prompt}
\begin{tabular}{p{0.95\linewidth}}
\toprule
\textbf{QA rating prompt} \\ 
\midrule
\begin{minipage}[t]{\linewidth}

You will evaluate two answers to the same question based on three criteria: \textbf{Comprehensiveness}, \textbf{Diversity}, and \textbf{Empowerment}. \\

- \textbf{Comprehensiveness}: How much detail does the answer provide to cover all aspects and details of the question?  \\
- \textbf{Diversity}: How varied and rich is the answer in providing different perspectives and insights on the question?    \\
- \textbf{Empowerment}: How well does the answer help the reader understand and make informed judgments about the topic? \\

For each criterion, choose the better answer (either Answer 1 or Answer 2) and explain why. Then, select an overall winner based on these three categories. \\

Here is the question: \{\texttt{query}\}    \\

Here are the two answers: \\
\textbf{Answer 1}: \{\texttt{answer1}\}  \\
\textbf{Answer 2:} \{\texttt{answer2}\}  \\

\vspace{1em}

Evaluate both answers using the three criteria listed above and provide detailed explanations for each criterion. And you need to be very fair and have no bias towards the order.  \\

Output your evaluation in the following JSON format:    \\
\begin{verbatim}
{{
    ``Comprehensiveness'': {{
        ``Winner'': ``[Answer 1 or Answer 2]'',
        ``Explanation'': ``[Provide explanation here]''
    }},
    ``Empowerment'': {{
        ``Winner'': ``[Answer 1 or Answer 2]'',
        ``Explanation'': ``[Provide explanation here]''
    }},
    ``Diversity'': {{
        ``Winner'': ``[Answer 1 or Answer 2]'',
        ``Explanation'': ``[Provide explanation here]''
    }},
    ``Overall Winner'': {{
        ``Winner'': ``[Answer 1 or Answer 2]'',
        ``Explanation'': ``[Summarize why this answer is the overall winner based on the 
                            three criteria]''
    }}
}}
\end{verbatim}

\end{minipage} \\

\bottomrule 
\end{tabular}
\end{table*}

\end{document}